\documentclass[final]{cvpr}

\usepackage{times}
\usepackage{epsfig}
\usepackage{graphicx}
\usepackage{amsmath}
\usepackage{amssymb}
\usepackage{multirow}
\usepackage[ruled,vlined]{algorithm2e}
\usepackage{dsfont}
\usepackage{xcolor}
\usepackage{caption}
\usepackage{footnote}
\usepackage{float}

\newcommand{\ZE}{\textcolor{blue}}

\usepackage[pagebackref=true,breaklinks=true,colorlinks,bookmarks=false]{hyperref}

\def\cvprPaperID{4074} 
\def\confYear{CVPR 2021}

\begin{document}
\title{Using Text to Teach Image Retrieval}



\author{Haoyu Dong \\
Duke University \\
{\tt\small haoyu.dong151@duke.edu}
\and
Ze Wang \qquad Qiang Qiu\\
Purdue University\\
{\tt\small \{wang5026, qqiu\}@purdue.edu}
\and
Guillermo Sapiro\\
Duke University\\
{\tt\small guillermo.sapiro@duke.edu}
}


\twocolumn[{%
\renewcommand\twocolumn[1][]{#1}%
\maketitle
\begin{center}
    \centering
    \includegraphics[scale=0.45]{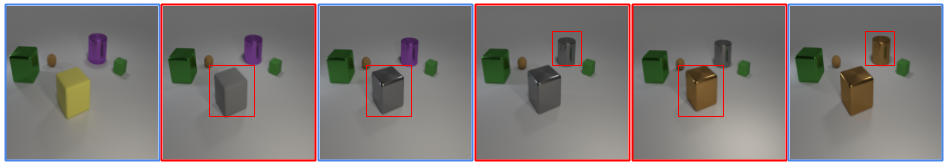}
    \captionof{figure}{Visualization of a smooth path as obtained with the framework proposed in this paper. Images with blue boundaries are from the image domain, and ones with red boundaries are from the text domain. The modification between adjacent images is marked with a red square, corresponding to a single attribute change per step on the joint embedding geodesic path. Best viewed in color.}
    \label{fig:patheval}
\end{center}%
}]

\begin{abstract}
    {
}

   Image retrieval relies heavily on the quality of the data modeling and the distance measurement in the feature space. 
   Building on the concept of \textit{image manifold}, we first propose to represent the feature space of images, learned via neural networks, as a graph. 
   Neighborhoods in the feature space are now deﬁned by the geodesic distance between images, represented as graph vertices or manifold samples. 
   When limited images are available, this manifold is sparsely sampled, making the geodesic computation and the corresponding retrieval harder. 
   To address this, we augment the manifold samples with geometrically aligned text, thereby using a plethora of sentences to teach us about images. In addition to extensive results on standard datasets illustrating the power of text to help in image retrieval, a new public dataset based on CLEVR is introduced to quantify the semantic similarity between visual data and text data. 
   The experimental results show that the joint embedding manifold is a robust representation, allowing it to be a better basis to perform image retrieval given only an image and a textual instruction on the desired modiﬁcations over the image.
   
   
\end{abstract}

\section{Introduction}

Retrieval is the task of finding the most relevant object in a database given a query. Recent works have grown the interest in cross-domain retrieval, especially between image and text domains. The image-text retrieval task can be generally summarized into two main directions.
One is to match the corresponding images given sentence queries, or vice versa; this has been used as an evaluation metric in many works that learn visual-semantic embeddings  \cite{faghri2018vse++, wang2019twobranch}. 
The other direction, \eg, \cite{Han2017AutomaticSF,vo2019composing}, is to conduct text-based retrieval; the task uses an image with a textual instruction describing some desired modifications to the image as a query. 
In both scenarios, the search is done by mapping the queries and database objects to a feature space. 

Although remarkable progress has been achieved, the basic frameworks of retrieval are mostly built upon the assumption that the information in the embedding space is well approximated by either negative Euclidean distance or negative cosine distance, both assuming that the features are in a Hilbert space. 
This can be sub-optimal under the common assumption that images reside on a low-dimensional manifold within a high-dimensional feature space \cite{10.1023/B:MACH.0000033120.25363.1e}, where a geodesic distance can better define the relationship between objects. 
Furthermore, since we usually only have access to a limited number of samples in the visual domain, the feature space is under sampled, and thus sparse \cite{viewinterpolation}. 
A sparse feature space means that some points can be far away from all the other points, making it hard to define proper neighborhoods for them, and to compute accurate geodesic distances.
These points are considered unretrievable, and our experiments show that they introduce most errors in the standard retrieval methods. 

In this work, we model the manifold of image features, learned via neural networks, as a graph, where each vertex represents an image. 
Considering that this manifold is only locally homeomorphic to Euclidean space, we build an edge between a pair of vertices only when their Euclidean distance is small, as standard in the point clouds literature.
Given this graph representation, we can now approximate the geodesic distance between a vertex and its neighborhood \cite{tenenbaum_global_2000}. 
Since edges represent distances, our image graph is weighted, and we compute the distance as the sum of weights along the shortest path instead of just as the number of edges.
To evaluate the quality of the distance measurement in the feature space, we study a label retrieval task which aims at classifying every node in a graph with labels only available for a small subset. 
The motivation behind this task is from semi-supervised learning, where the goal is to propagate label information in a naturally defined fashion.
This learning task has a key assumption, that points in the same locality are likely to share the same label \cite{NIPS2002_2257, Zhou2003LearningWL}. 
Therefore, the task can be seen as a natural way to measure the robustness of the feature representation.

The geodesic distance and manifold concept further allow us to consider the sparsity problem, meaning the limited number of samples (vertices), in the retrieval task. 
When the number of samples is limited, some samples are far from the rest.
Then the degree of these samples are small, meaning they only have a few or even do not have geodesic neighbors.
We say a point is retrievable if it has a geodesic neighbor in the small subset that contains label information, and otherwise it is unretrievable.
We propose to exploit plethora of text in order to learn a visual-semantic embedding space to reduce the number of unretrievable points and to improve the geodesic computation accuracy \cite{tenenbaum_global_2000}, see Fig. \ref{fig:patheval}.
The objective of a joint embedding space is to improve the learned image manifold representation by adding, to the original image samples (vertices), new manifold samples via semantically related text. 
We further evaluate the manifold in the joint embedding space by counting the total number of ``smooth" shortest paths starting from all images points. 
The intuition is that if text samples are drawn from the same distribution (manifold) as the image samples, they can be interpolated among image embeddings and the transitions among text and image samples are smooth and meaningful. 
To quantitatively define the concept of ``smoothness", we use the CLEVR framework \cite{johnson2017clevr} and introduce a new dataset, named ``CCI" (CLEVR-Change-Iterative).
The framework provides a set of objects with different colors, shapes, materials, etc., and we consider a transition to be smooth if both images differ by only one attribute.
A shortest path is then smooth if all the transitions along the path are smooth. 
We can therefore study how well the textual domain is incorporated into the visual domain by examining the increase in the number of smooth shortest paths.

{
}

We empirically conduct label retrieval experiments on two natural image sets: ADE20K \cite{zhou2017scene} and OpenImage \cite{OpenImages}. 
We show that a geodesic neighbor leads to a better retrieval performance than an Euclidean neighbor does. 
Furthermore, we observe that using image and textual information together to represent the manifold allows a more completed neighborhood description, where more points are retrievable.
To validate the proposed alignment of the text domain with the image domain, we build a cross-modal embedding space with a ranking loss on CCI. We compare our learned text embeddings with random generated embeddings to show that merely having more samples does not increase as many smooth paths as semantically similar embeddings do. 

Interestingly, we also find that the manifold in the joint feature space, with only image samples, outperforms the one in a pretrained image feature space. 
This means that we can use text as
information to learn a more robust representation of images. 
To validate this statement, we use the collection of new image features as a basis for text-based retrieval. 
We obverse consistent improvements over different embedding methods and different datasets.  
We also show that our new dataset, CCI, can be used to train text-based retrieval. 

{
}

In conclusion, our contribution are three-fold.
\begin{itemize}
    \item We demonstrate that the geodesic distance is a more accurate measurement to the relationship between objects. Adding corresponding texts can alleviate the sparsity problem, and improve the embedding manifold representation. 
    \item We show that corresponding texts can also be used as privileged information for text-based retrieval.
    \item We introduce a new public dataset, CCI, that can be used to validate the quality of a joint feature embedding space, as well as to do text-based retrieval task. 
\end{itemize}


\section{Prior Work} 
\noindent \textbf{Fusion of Vision and Language.} There is a growing interest to solve the problems at the intersection of computer vision and natural language processing; the problems range 
from transferring information from one domain to another, \ie, image captioning \cite{Anderson2018BottomUpAT, Gan2017SemanticCN, Xu2015ShowAA} and text-to-image generation \cite{Hinz2020SemanticOA, pmlr-v48-reed16},
to integrating information from both domains to solve questions that require cross-modal knowledge, such as visual question answering (VQA) \cite{Agrawal2015VQAVQ, Kazemi2017ShowAA, Tan2019LXMERTLC} and novel object captioning \cite{Hu2020VIVOSH, Lu2018NeuralBT, Yao2017IncorporatingCM}.
These cross-domain problems usually require a task-specific framework to answer, although some works try to learn a visual-semantic joint embedding space that can be more adaptive to multiple tasks \cite{faghri2018vse++, wang2019twobranch}.
Of particular interest, we find that cross-domain learning brings alignment that can be viewed as an unsupervised way to solve problems. 
For example, \cite{Harwath2018JointlyDV} finds that regions which fire high in spatial activation maps correspond to the relevant objects described in the speech; \cite{sigurdsson2020visual} builds an unsupervised translation system with the assumption that sentences in different languages appearing with visually similar frames are correlated. 
In both cases, specific relations, \ie, segments of objects and parallel corpora, are not available. 
In our work, we find that captions (text) can be seen as extra samples from the same distribution (manifold) to provide a richer representation of the image space helping in label retrieval, even though the label for each sentence is unknown. 


\noindent \textbf{Cross Domain Retrieval.} Image retrieval that relies on deep learning approaches usually encodes images by Convolutional Neural Networks (CNNs) that are pretrained on classification problems, and has gained significant progress \cite{Gordo2016DeepIR, Wang2014LearningFI}. 
Cross-domain retrieval extends the problem by considering queries that are more than an image, such as text to image retrieval \cite{Wang2016LearningDS} and sketch to image retrieval \cite{Sangkloy2016TheSD}.
The works \cite{Han2017AutomaticSF, vo2019composing} further study text-based retrieval, where each query is composed of an image plus some instructions describing the desired modifications to that image; the framework allows users to provide feedback to product search.
Based on our observation that image samples are more robustly represented in the joint space, and the task's tendency to incorporate modification sentences into an image representation, we compose text embeddings with image embeddings from the joint space instead, and observe that this outperforms composing with image embeddings from the pretrained embedding space.
Parallel to our work, \cite{chen2020Joint} has incorporated side information into the text-based retrieval task. Though the approach is similar, we see the improvement as a proof of concept. 

\noindent \textbf{Semi-Supervised Learning.} 
Semi-supervised learning has two key assumptions of consistency, which are: (1) nearby points are likely to share the same label; and (2) points with similar structure are likely to have the same label \cite{Zhou2003LearningWL}.
Recent works that use graph convolutional networks (GCN) follow the second assumption and achieve promising results \cite{Kipf2017SemiSupervisedCW}. 
Although we also represent image features as a graph, the graph is used to calculate shortest path distances and to propagate label information. 
Our approach is still $K$-Nearest Neighbor (KNN) based, which follows the first assumption. 
We thus believe the improvement in label retrieval performance reflects that points with the same label are closer, \ie, the representations are more robust.

\begin{figure}[h]
    \centering
    \includegraphics[scale=0.25]{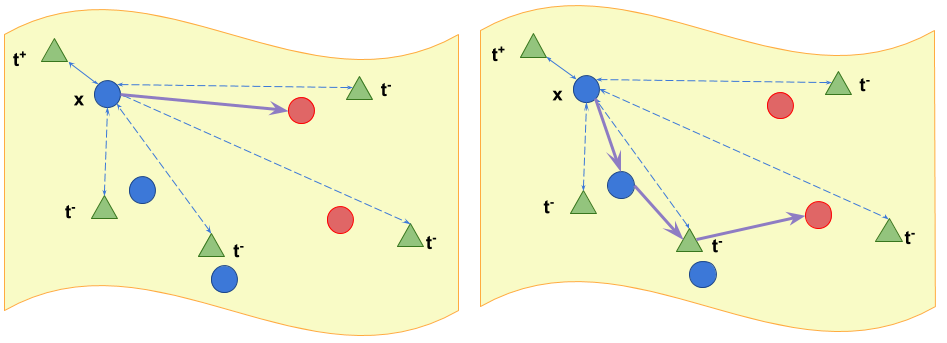}
    \caption{Illustration of learning the cross-domain feature space and two corresponding distances. An image point (blue circle) is encouraged to be close to its corresponding text point (green triangle) $t^+$, and far from non corresponding text points $t^-$. Left and right figures show finding the nearest point in the target set (red circles) by ranking Euclidean and geodesic distances, respectively. Best viewed in color.}
    \label{fig:illustration}
\end{figure}

\section{Method}
In this section, we first describe the proposed method of learning a visual-semantic joint embedding by incorporating text information into the visual domain. 
Then we present the approach to construct a graph in the joint space. 
Finally, we show how to use image features from the visual-semantic domain as a basis to conduct text-based retrieval. Figure \ref{fig:illustration} illustrates the proposed architecture. 


\subsection{Cross-Modal Embedding}
To learn the manifold and correspondence between images and texts in a joint feature space, we use a two-branch neural network similar to \cite{wang2019twobranch, Zhang2018DeepCP}.

\noindent \textbf{Visual Embedding Module}. We use a CNN to project the input images $x$ to image embeddings, $f_{img}(x) = \psi_x \in \mathds{R}^d$. In our experiments, we use ResNet-18 \cite{He2016DeepRL} and replace the classification layer with a fully-connected layer with $d$ units. 

\noindent \textbf{Textual Embedding Module}. We encode the descriptive texts $t$ using a Universal Sentence Encoder (USE) \cite{2018arXiv180311175C}. The encoded features are passed through two fully-connected layers with ReLU activation, $f_{text}(t) = \phi_t \in \mathds{R}^d$. In our experiments, the fully-connected layers are learned from scratch.  

\noindent \textbf{Semantic Projection Layers}. The projection layers map both the image and text embeddings onto a joint latent space in which matched image and text features are pushed closer, while unmatched ones are pulled apart. We add a fully-connected layer after each module, followed by L2 normalization, denoted as $g$. Suppose we have a mini-batch of $B$ images and texts pairs, $\{(x_i, t_i)\}_{i=1}^B$, we encode each image $x_i$ as $\psi_i = g(f_{img}(x_i))$, its corresponding text $t_i$ as $\phi_i^+ = g(f_{text}(t_i))$, and the rest of the text in the mini-batch as $\phi_{j,j\ne i} = g(f_{text}(t_j))$. Then, we define the objective as

\newcommand*{\Scale}[2][4]{\scalebox{#1}{$#2$}}%
\begin{equation}
\label{equation:rank}
\Scale[1.1]{
    L = \frac{1}{B}\sum_{i=1}^{B} -log \large\{ {\frac{ \text{exp}(\kappa(\psi_{i}, \phi_{i}^+))}{\sum_{j=1}^{B} \text{exp} (\kappa(\psi_i, \phi_j))}} \large\},
}
\end{equation}

\noindent where $\kappa(\cdot, \cdot)$ is a kernel function implemented as a dot product, which is equivalent to the Euclidean distance after normalization. This loss resembles a cross-entropy loss to classify which ``text class" the image belongs to \cite{gidaris2018dynamic, NIPS2004_2566,MovshovitzAttias2017NoFD}.

\subsection{Graph in the Joint Space}
\label{section:classpred}
We now explain in details how to construct a graph in the joint feature space. 
In order to promote better feature correspondence between image embeddings and text embeddings, we perform feature alignments to reduce the distance between pairs. Specifically, we adopt the iterative point alignment (ICP) algorithm \cite{Besl1992AMF}.
After alignment, we compute local edges using the great-circle distance instead of the L2 distance to better mimic a ``walk" on the manifold. We describe these steps next. 

\noindent\textbf{Point Alignment}.
Given all image and text embeddings, $(\Psi, \Phi)$, we further align the pairs using ICP transformation before constructing the graph. Since nearest points are given as the corresponding pairs, the algorithm only needs to be run once as described in Algorithm \ref{algo:icp}. 
\begin{algorithm}
\SetAlgoLined
$\hat{\Psi} = \Psi - mean(\Psi), \hat{\Phi} = \Phi - mean(\Phi)$\;
$U, S, V_t = SVD(\hat{\Psi}^T * \hat{\Phi})$\;
$R = V_t^T*U^T$\;
$t = mean(\Psi) - mean(\Phi)$\;
$T[:d,:d] = R, T[:d,d] = t$\;
\caption{ICP}
\label{algo:icp}
\end{algorithm}

\noindent\textbf{Graph Construction}.
Given the aligned embeddings of image and text, $(T(\Psi), \Phi)$, we can now construct the graph G.
The nodes (vertices) are either an image embedding or a text embedding, and edges encode distances between two points. Since we map all the points to the joint space with L2 normalization, we use the great-circle distance on the unit $n$-sphere. 
Following the locally-Euclidean property of a manifold, an edge $(i,j)$ exists \textit{iff} the distance between $i$ and $j$ is lower than a given threshold, which is an hyperparameter to define locality and usually set to satisfy that $O(|E|) = O(|V|)$. 





\subsubsection{Manifold Evaluation}
To evaluate the quality of neighbors and the shortest paths, we introduce two tasks, label retrieval and smooth path counting.

\noindent\textbf{Label Retrieval.}
The task first selects a small subset of images as the database (target set), where label information is available.
The unselected images serve as the queries, whose labels are predicted by KNN with geodesic neighbors. 
When more samples from the text domain are presented, they are used as privileged information, and thus their own label information is not computed.


\noindent\textbf{Smooth Path Counting.} 
In this task, we count the number of smooth paths in the graph $G$. The ``smoothness" is defined in Sec. \ref{subsection:patheval}, and 
we run Dijkstra's algorithm on every image vertex, and count the shortest paths that end with another image vertex. 
In this way, only paths starting and ending in the image domain are counted, so that text vertices can be viewed as additional guidances that connect pairs of related images. 


\subsection{Retrieval in the Joint Space}
\label{section:textbasedretrieval}
Text-based retrieval is similar to standard image retrieval task, except that we need an additional module, $\psi_{xt} = f_{compose}(\psi_x, \phi_t)$, to compose image and text features. 
The composition of images and texts can be seen as moving the image features along the direction learned from the texts;
we propose to perform moving image features in the joint space to improve performance, based on the observation that the representation of images is more robust in this space. We detail this process next.


\noindent\textbf{Compositional Module.}
There are existing methods to compose $\psi_x$ and $\phi_t$. 
We study one explicit method for the text-based retrieval task and one universal method to learn the general relationship among entities in the embedding space. 
The objective is to enforce that the distances between the composed features and the targets are shorter than the distances between unmatched pairs. 
We can also use Eq. (\ref{equation:rank}) to train the compositional module. 
The two methods are described as follows.
\begin{itemize}
    \item \textbf{TIRG} \cite{vo2019composing} learns gating features and residual features from images and queries, and the output is the weighted sum of the two features. 
    The residual features are learned by applying two 3x3 convolution layers with non-linearity on the concatenation of $[\psi_x, \phi_t]$.
    The gating function uses the same module followed by a Sigmoid function to learn a ``filter" for the image features, \ie, the output is the element-wise product of the filter signal and $\psi_x$.
    \item \textbf{Relationship} \cite{Santoro2017ASN} is a VQA method that captures the relational reasoning. It forms a relational set by combining 2 local features $\psi_x(i,j)$ and $ \psi_x(i',j')$ with text features $\phi_t$. The set is passed through multi-layer perceptrons (MLPs), and the sum of the output is passed through another MLPs to get the final output.
\end{itemize}

\noindent\textbf{Retrieval in the Joint Space.} To compose texts and images in a joint feature space, we can simply insert the projection layer as a plug-and-play module to the compositional functions. Denote the target images as $x^{target}$ and the modification queries as $t^{query}$, the loss becomes
\begin{equation}
    L = L(g(\psi_x), g(\phi_t)) + L(\psi_{xt}, g(\psi_x^{target})),
\end{equation}
where $\psi_{xt} = f_{compose}(g(\psi_x), g(\phi_t^{query}))$ is the composed feature. In our experiments, we simplify the text encoder to a standard LSTM in order to introduce minimum modifications over the original method. 
An identical text encoder is trained to encode the query texts separately.

\section{Experimental Protocol and Results}
To quantitatively evaluate the performance of our manifold-based retrieval, we present the results on label retrieval tasks and text-based retrieval tasks. 
In both tasks, the evaluation metric is the recall at rank $K $($R$@$K$), as the percentage of test queries where the target image is within the top-$K$ retrieval samples. 
For label retrieval tasks, we only consider $R$@$1$ because it resembles an image classification evaluation metric. 

We use PyTorch in our experiments. The image encoder is the ResNet-18 pretrained on Imagenet throughout all experiments. 
When constructing the graph, we use the latest USE (V4) in tensorflow-hub (embedding size 512), and the outputs are converted to PyTorch tensors. We adopt SGD optimizer with a starting learning rate of 0.01 with batch size of 50 for 200k iterations. 


\begin{figure*}[ht]
    \centering
    \includegraphics[scale=0.45]{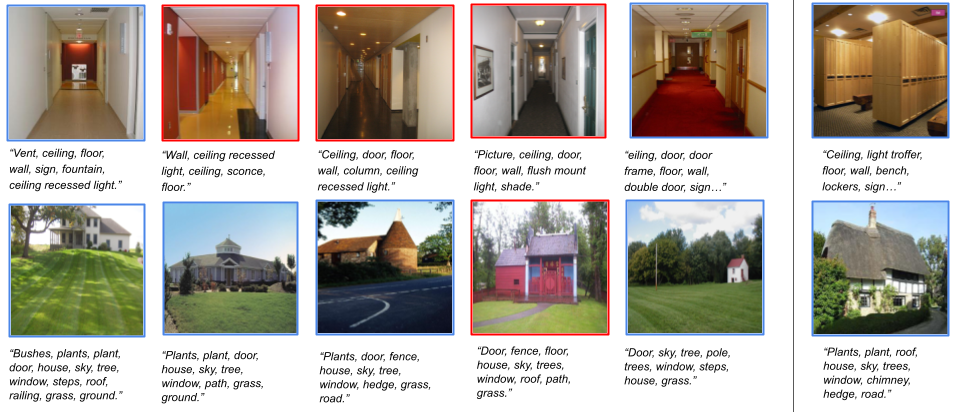}
    \includegraphics[scale=0.45]{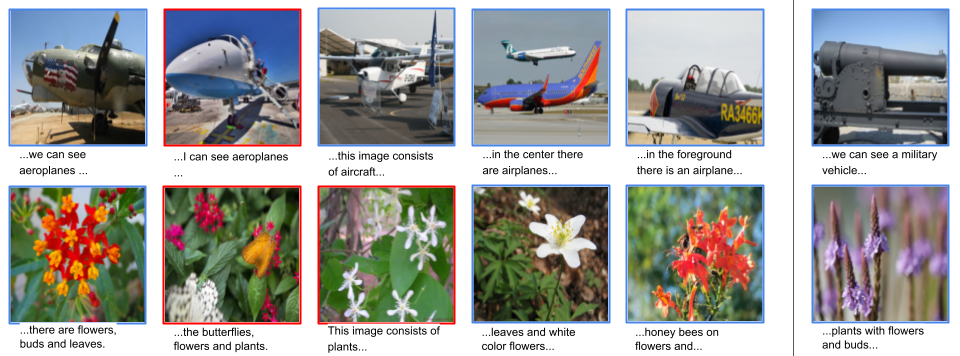}
    \caption{Qualitative examples from ADE20K (top 2 rows) and OpenImage (bottom 2 rows). We present each manifold point with its original input (an image or a sentence) and its corresponding point from the other domain. 
    Image points are presented with blue boundaries, and text points' corresponding images are presented with red boundaries. 
    Column 1 is the starting vertex (always from the image domain), and column 5 is its nearest geodesic neighbor, with columns 2-4 presenting the shortest path. Column 6 is the nearest Euclidean neighbor.}
    \label{fig:classcorrect}
\end{figure*}

\subsection{Datasets}
\label{subsection:dataset}
\noindent\textbf{Class Prediction}. 
ADE20K \cite{zhou2017scene} contains a wide range of objects in a variety of contexts. To ensure that each image has a precise label, we exclude images from the ``Outlier" and ``MISC" categories. This leads to 715 classes, and 17,956 images. 
There are no corresponding captions for each image, but we can extract side information from the segmentation at the object level. 
OpenImage \cite{OpenImages} is a very large dataset with rich information. We only use captions for training and labels for testing. We train under the validation set because it has more precise side information, \eg, human-verified labels, and the original training set ($\sim 9M$ images) is beyond our scope. In total, there are 528 classes, and 41,620 images. Each image is annotated with multiple labels, and we consider the classification is correct \textit{iff} all labels are matched. 

\noindent \textbf{Text-based Retrieval.} We use Fashion-IQ \cite{guo2019fashion}, CSS \cite{vo2019composing}, and a new synthetic dataset we named CCI (see Section \ref{subsection:newdataset}). These datasets contain attribute-like descriptions for images, and textual modification instructions between pairs of images. Fashion-IQ contains 18,000 training samples.
We use all the pairs that the side information for both images is available, which leads to 8,847 samples. We evaluate the performance on the validation set.\footnote{We evaluate on the val set since the ground truth for test set is not released.} 
CSS and CCI are two CLEVR-based datasets, where each image contains attribute information for all objects within the image, and the instructions are generated from templates. We follow the standard train-test splits \cite{vo2019composing} for CSS. Specifically, CSS has 18,012 training samples and 18,057 testing samples; CCI has 1,110 training samples and 10,000 testing samples. 


\subsection{CLEVR-Change-Iterative Dataset}
\label{subsection:newdataset}
Existing benchmarks focus on the scale and diversity of images. 
However, to provide a measurable definition of smoothness, it is more desirable to let every image be a variance of a ``source" image, so that the difference between any two images can be categorical and thus countable.
Specifically, we use the CLEVR toolkit \cite{johnson2017clevr} to create our dataset. First, we generate one scene with random objects to serve as the ``source" image. 
For this image we create ten ``modified" images and the corresponding modification instructions following \cite{park2019robust}. In particular, we apply \textit{camera position changes} and \textit{scene changes} besides DISTRACTOR to ensure every image is unique. 
We iterate the modification process by considering the ten "modified" images as the ``source" ones, and applying the same modification process on each of them, respectively. 
In total, we repeat the process four times, and generate 11,111 images.

To quantify the semantic similarity between images and texts, we provide neighbor information for each image (see Sec \ref{subsection:patheval} for details). 
Specifically, we define every image's reachable neighbors in the following way: (a) Two images only differ by a single attribute of a single object; or (b) Two images only differ by one having a single additional object.
On average an image has 9.73 reachable neighbors. 
All images are used for the smooth path counting task.

To use the dataset for the text-based retrieval task, we consider each modification providing a pair of source and target images and the corresponding modification instructions.
We use the pairs from the first three iterations as the training set, and the ones from the last iteration as the target set. This results in 1,110 training samples and 10,000 testing samples.

\subsection{Label Retrieval}
\label{subsection:classretrieval}
The database (target set) we adopt for label retrieval is constructed in an $N$-way-$k$-shot fashion.
In this way, the label information is evenly distributed in the manifold, and all images from under-presented classes, \ie, classes with less than $k$ samples, are excluded from the label retrieval tasks.
Since OpenImage \cite{OpenImages} contains multiple labels per image, we use each image's most significant label, detected by the corresponding bounding box, to assign the classes. 
However, we do not only use these single labels for the task because matching all labels of an image can be a more faithful measurement of locality;
experiments in addition show that single-label retrieval leads to a lower recall score than multi-labels retrieval does, which suggests that these single labels are not well-aligned with the images.

We compare Euclidean neighbors and geodesic neighbors, denoted as \textit{Eu} and \textit{Geo} for short. 
If an image does not have a single Euclidean neighbor in the target set, this image point is considered unretrievable; otherwise it is retrievable. 
Our method is conducted on three feature spaces, $T(\Psi), T(\Psi)+\Phi$, and $\Psi'$ from the pretrained ResNet, named as \textit{Image}, \textit{Joint}, and \textit{Resnet}.
We do not consider $\Psi + \Phi$, \ie, the joint feature space without applying ICP, because experiments show that without the ICP transformation, there will be no cross-domain edges under the locality threshold. 
Note that the comparison between $T(\Psi)$ and $\Psi'$ is appropriate, since ICP is a rigid transformation that preserves the distances among vertices. 

\begin{figure*}[h]
    \centering
    \includegraphics[scale=0.45]{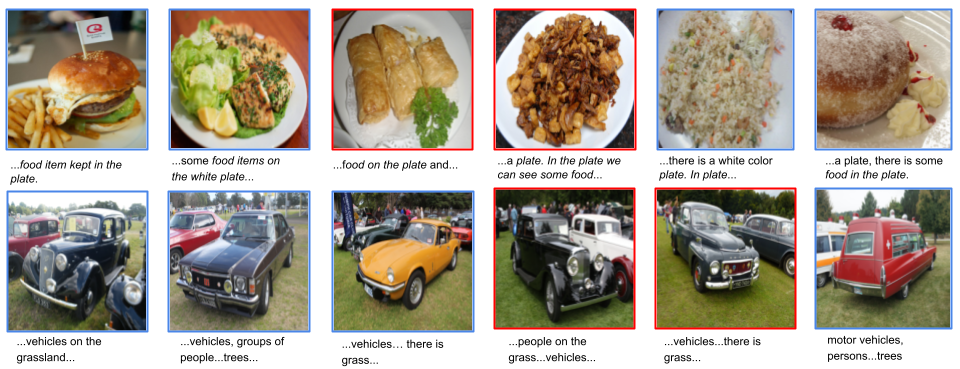}
    \caption{Qualitative failure examples from OpenImage. Column 6 is now the nearest geodesic neighbor, with columns 2-5 presenting the shortest path. Row 1 shares a general concept, ``foods on a plate," and row 2 shares a general concept, ``cars on grass."}
    \label{fig:classincorrect}
\end{figure*}

\begin{table*}[ht]
    \centering
    \begin{tabular}{|c|c|c|c|c|c|}
     \hline
     & \multirow{2}{*}{Method} & \multicolumn{2}{c|}{ADE20K} & \multicolumn{2}{c|}{OpenImage} \\
     & & Accuracy & Retrievable Points & Accuracy & Retrievable Points \\
     \hline
     \hline
     A & Baseline (\textit{Image + Eu}) & 0.1415 & 15925 & 0.0609 & 39665\\
     \hline
     B &\textit{Resnet + Geo} & 0.1524 & 4169 & 0.1199 & 5569\\
     C & Baseline$^*$ in \textit{Image} & 0.2056 & 4169 & 0.1270 & 5569 \\
     D &\textit{Image + Geo} & \textbf{0.2770} & 4169 & \textbf{0.1356} & 5569 \\
     E & Baseline$^*$ in \textit{Joint} & 0.2087 & 5774 & 0.1198 & 7461 \\
     F &\textbf{\textit{Joint + Geo}} & \underline{0.2703} & \textbf{5774} & \underline{0.1328} &\textbf{7461} \\
     \hline
    \end{tabular}
    \caption{Quantitative results of label prediction on ADE20K and OpenImage. Baseline$^*$ indicates applying \textit{Eu} method on retrievable points. Highest accuracy is highlighted and second highest accuracy is underlined.}
    \label{tab:classificationade}
\end{table*}

\subsubsection{Results} 
\label{subsectin:classpredresult}
Tables \ref{tab:classificationade} summarizes the quantitative results on ADE20K and OpenImages. 
Our baseline shows the performance of the typical retrieval approach (by ranking Euclidean distances).
To make it comparable to a manifold-based approach, we run the baseline again only on the retrievable points.  
The comparisons (C vs.\ D and E vs.\ F) reflect that exploiting the geodesic-based nearest neighbors outperforms the Euclidean-based nearest neighbors. 
When considering the baseline on retrievable points versus all points (A vs.\ C or E), we see the performance increases when only computing on retrievable points (that are closer to other points).
This validates our statement that the image points far from the database introduce more errors in the retrieval task. 
Note that the same mechanism works better on ADE20K; the reason could be that ADE20K provides side information at the object level, whereas OpenImage gives natural language captions that are likely to describe images at a higher level.  
When text points are introduced as additional samples to represent the manifold, the manifold has more retrievable points, while the accuracy ($R$@$1$) remains similar (D vs.\ F). 
The improvement suggests that the originally sparse visual feature space is now interpolated by the dense knowledge from the textual domain that is properly aligned with the visual domain.

Finally, we compare the manifolds constructed with image features from \textit{Image} or \textit{Resnet} (B vs.\ D). 
Since now features come from different embedding spaces, we approximate the manifold from \textit{Resnet} that gives the same amount of retrievable points by finding two manifolds, one having slightly more retrievable points, and the other having slightly fewer, and interpolate between them. 
We can see the image features from the joint space give better representation than from the pretrained space, and this motivates our experiment in Sec. \ref{subsection:textbasedretrieval}.

\begin{table*}[ht]
\centering
\begin{tabular}{|c|cc|cc|cc|}
    \hline
    \multirow{2}{*}{Method} & \multicolumn{2}{c|}{Dress} & \multicolumn{2}{c|}{Toptee} & \multicolumn{2}{c|}{Shirt} \\
    & $R$@$10$ & $R$@$50$ & $R$@$10$ & $R$@$50$ & $R$@$10$ & $R$@$50$ \\
    \hline
    \hline
    TIRG \cite{vo2019composing} & 0.1264 & 0.3386 & 0.1545 & 0.4080 & 0.1457 & 0.3690 \\
    TIRG +\textit{ in Joint} & \textbf{0.1507} & \textbf{0.3644} & \textbf{0.1856} & \textbf{0.4518} & \textbf{0.1638} & \textbf{0.4028} \\
    \hline
    Relation & 0.0744 & 0.2137 & 0.0918 & 0.2478 & 0.1084 & 0.2561 \\
    Relation +\textit{ in Joint} & \textbf{0.0843} & \textbf{0.2568} & \textbf{0.1147} & \textbf{0.2917} & \textbf{0.1178} & \textbf{0.2870} \\
    \hline
\end{tabular}
\caption{Retrieval Performance ($R$@$10$ and $R$@$50$) on Fashion IQ. The recall scores are higher when the method runs on the joint space. All results are from our implementation.}
\label{table:fashioniqretrieval}
\end{table*}

\begin{table}[ht]
    \centering
    \begin{tabular}{|c|c|c|c|}
        \hline
        Threshold & $\Psi$ & $\Psi +$ Random & $\Psi + \Phi$ \\
        \hline
        \hline
        0.50 & 9.96 & 10.49 & 10.78 \\
        0.52 & 10.29 & 10.83 & 11.20 \\
        0.54 & 10.60 & 11.18 & 11.66 \\
        0.56 & 10.95 & 11.56 & 12.11 \\
        0.58 & 11.31 & 11.82 & 12.54 \\
        \hline
    \end{tabular}
    \caption{Log of number of paths under different thresholds and feature space. $\Psi$ stands for image features; $\Phi$ stands for text features; and random stands for random features. Note that under different thresholds adding text features leads to more smooth paths than adding random features.}
    \label{tab:path}
\end{table}

\subsubsection{Path Evaluation}
\label{subsection:patheval}
We present some qualitative results of geodesic and Euclidean neighbors in 
Fig. \ref{fig:classcorrect}, which also illustrates the shortest paths to the geodesic neighbors. 
We observe that traversing the graph with small steps allows for an image to find other images in the same neighborhood, while an Euclidean path may lead to an image from a different neighborhood. 
We include some qualitative results for unsuccessful prediction cases on ADE20K in Fig. \ref{fig:classincorrect} as well. 
The first example illustrates that nearby features from a visual-semantic joint space can sometimes be semantically similar but visually different. 
The second example reflects that the label retrieval task is only a loose representation for the robustness of feature spaces. Specifically, the starting image is labeled as ``car, land vehicle" and the nearest geodesic neighbor is labeled as ``land vehicle, limousine, van". The inconsistency in labeling introduces additional errors.

In both successful and unsuccessful examples, we consistently observe that the transitions between image vertices and text vertices are smooth.
This illustrates that the additional vertices from the text domain are well-aligned with the original image vertices.
To quantitatively measure the alignment of a joint space, we introduce our new CCI dataset to handcraft a criteria for the ``smoothness" of shortest paths.
We first define a transition between two images to be smooth if they are in each other's reachable neighbors. 
We further restrict the path to be non-redundant, \ie, for any vertex in the path, its non-adjacent vertices cannot be reached by it, otherwise the paths between the two vertices can be ignored.
Considering these two concepts, we can define a path to be smooth \textit{iff} the transitions between every two adjacent vertices are smooth and the transitions between any two non-adjacent vertices are not smooth. 
An example of a smooth path is shown in Fig. \ref{fig:patheval}. 

Then we can count the number of smooth shortest paths to reflect the symmetry between image and text features.  
The result is shown in Table \ref{tab:path}. Our baseline is to construct the manifold with image features only. When more samples are used to represent a manifold, we expect the number of valid paths to increase, and the level of increment reflects the semantic similarity. 
As finding an optimal method to learn a well-aligned joint space is beyond the scope of our paper, we only compare adding text features from the two-branch network with randomly generated features to show that having more semantically similar features results in more smooth paths. 

\subsection{Text-based Retrieval}
\label{subsection:textbasedretrieval}
As image features from a visual-semantic joint space are more robust, we can extend our work to text-based retrieval by adding the projection module (a fully-connected layer) after the encoders and one additional text encoder to encode side information.
We replace the language encoder with a standard LSTM (hidden size is 512) to allow for a fair comparison to \cite{vo2019composing}. 
Our method is denoted as \textit{in Joint} to emphasis that it only projects features into the cross-modal embedding without other modification to the original structure.
We also adopt the framework from \cite{vo2019composing} that allows different compositional methods .
The networks are trained from scratch, except for the TIRG method on the CSS dataset, where a pretrained model is available.  

Table \ref{table:fashioniqretrieval} shows R@10 and R@50 on the FashionIQ dataset under different categories, and 
Table \ref{table:cleverretrieval} summaries R@1 performance on the CSS and CCI datasets.
The performance of retrieval on the joint embedding space outperforms the one on the pretrained image space across different datasets. 
We note that the improvements are more significant on the CLEVR-based dataset than on FashionIQ. 
This is likely because the attributes on FashionIQ are less precise, \eg, a multi-color shirt can refer to different combination of colors, while a red object is definite. Moreover, we note there is a large margin in the performance on CCI. 
As any two images in this dataset are similar (since all images are derived from the same original image), it is more challenging for an image encoder to learn a discriminative feature space, where 
images are better separated.

\begin{table}[H]
\centering
\begin{tabular}{|c|cc|}
\hline
Method & CSS & CCI \\
\hline \hline
TIRG \cite{vo2019composing} & 0.7525 & 0.4123 \\
TIRG \textit{in Joint} & \textbf{0.7995}  & \textbf{0.7802} \\
\hline
Relation \cite{Santoro2017ASN} & 0.5301 & 0.2121 \\
Relation \textit{in Joint} & \textbf{0.5531} & \textbf{0.6125} \\
\hline
\end{tabular} 
\caption{Retrieval performance ($R$@$1$) on CSS and CCI. The recall scores are higher when the method runs on the joint space. All results are from our implementation.}
\label{table:cleverretrieval}
\end{table}

\section{Conclusion and Future Work}
We investigate the sub-optimal assumption that relation between images can be approximated by  negative Euclidean distance, and propose that a manifold structure and geodesic distances are a better representations. 
We further study the manifold of a joint embedding space, where text points can be used as additional samples.
These text samples are shown to benefit the image retrieval task, but we believe the application goes beyond this.
For example, texts can connect visually different but semantically similar images, which can be useful to learn a relational network; they can be further incorporated into GCNs as extra neighborhood information.
Overall, we show a way to represent the joint space of images and texts as a graph, opening multiple possibilities to interact with it.

\clearpage
{\small
\bibliographystyle{ieee_fullname}
\bibliography{egbib}
}

\end{document}